\documentclass[letterpaper, 10 pt, conference]{ieeeconf}  

\IEEEoverridecommandlockouts                              

\usepackage{helvet}         		
\usepackage{type1cm}        		
\usepackage{graphics} 		
\usepackage{graphicx}        
\usepackage[export]{adjustbox}
\usepackage{epsfig} 			
\usepackage{multicol}        		
\usepackage[hang,flushmargin]{footmisc}

\usepackage{amssymb}
\usepackage[ruled]{algorithm2e}
\usepackage{bm}
\usepackage{newtxtext}       %
\usepackage{newtxmath}       
\usepackage{caption}
\usepackage{gensymb}
\usepackage{graphicx}
\usepackage{booktabs}
\usepackage[table,xcdraw]{xcolor}
\usepackage{epstopdf}
\usepackage{url}
\usepackage[font={small}]{caption}
\usepackage{commath}
\usepackage{subfigure}
\usepackage{multirow}
\usepackage{makecell}
\usepackage{calc}
\usepackage{pifont} 
\usepackage{tikz}

\usepackage{algorithm2e}
\makeatletter
\renewcommand*{\@algocf@post@ruled}{}
\makeatother

\title{\LARGE \bf
Multi-Robot Autonomous Exploration and Mapping Under Localization Uncertainty with Expectation-Maximization
\vspace{-4mm}}

\author{Yewei Huang$^{1}$, Xi Lin$^{1}$ and Brendan Englot$^{1}$\vspace{-6mm}
\thanks{$^{1}$Y. Huang, X. Lin and B. Englot are with Stevens Institute of Technology, Hoboken, NJ, USA.
        {\tt\footnotesize \{yhuang85, xlin26, benglot\}@stevens.edu} \;\;\; This work was supported by ONR Grant N00014-21-1-2161.}
}
\begin{document}
\maketitle
\thispagestyle{empty}
\pagestyle{empty}

\begin{abstract}

We propose an autonomous exploration algorithm designed for decentralized multi-robot teams, which takes into account map and localization uncertainties of range-sensing mobile robots. Virtual landmarks are used to quantify the combined impact of process noise and sensor noise on map uncertainty. Additionally, we employ an iterative expectation-maximization inspired algorithm to assess the potential outcomes of both a local robot's and its neighbors' next-step actions.
To evaluate the effectiveness of our framework, we conduct a comparative analysis with state-of-the-art algorithms. 
The results of our experiments show the proposed algorithm's capacity to strike a balance between curbing map uncertainty and achieving efficient task allocation among robots.
\end{abstract}
\vspace{-2mm}
\section{Introduction}
Autonomous exploration and mapping describes a single robot or a group of robots navigating themselves in an unknown or partially known environment without human intervention.
An accurate environment map constructed by the robots serves as the fundamental basis for all subsequent specific robot tasks \cite{10075065}.
While the maturity of autonomous exploration for ground and aerial robots has been notably demonstrated, its application within marine environments remains a subject of ongoing inquiry \cite{cadena2016past}. 
In spite of the widespread coverage of Global Navigation Satellite System (GNSS) signals in most maritime regions, there exist many marine locations where GNSS signals are vulnerable due to obstruction or attenuation. 
In these areas, robots face a higher risk of collisions, emphasizing the critical need for an accurate environment map.
However, deploying exploration in such contexts is a particular challenge for robot teams. 
This is primarily due to the high uncertainty introduced by unique environmental factors pertaining to the marine environment, especially when operating below the ocean's surface.
\begin{figure}[t]
  \centering
  \subfigure[Choosing a revisiting frontier (blue, with red star) driven by the significant uncertainty in the map.]{%
    \includegraphics[width=0.45\columnwidth]{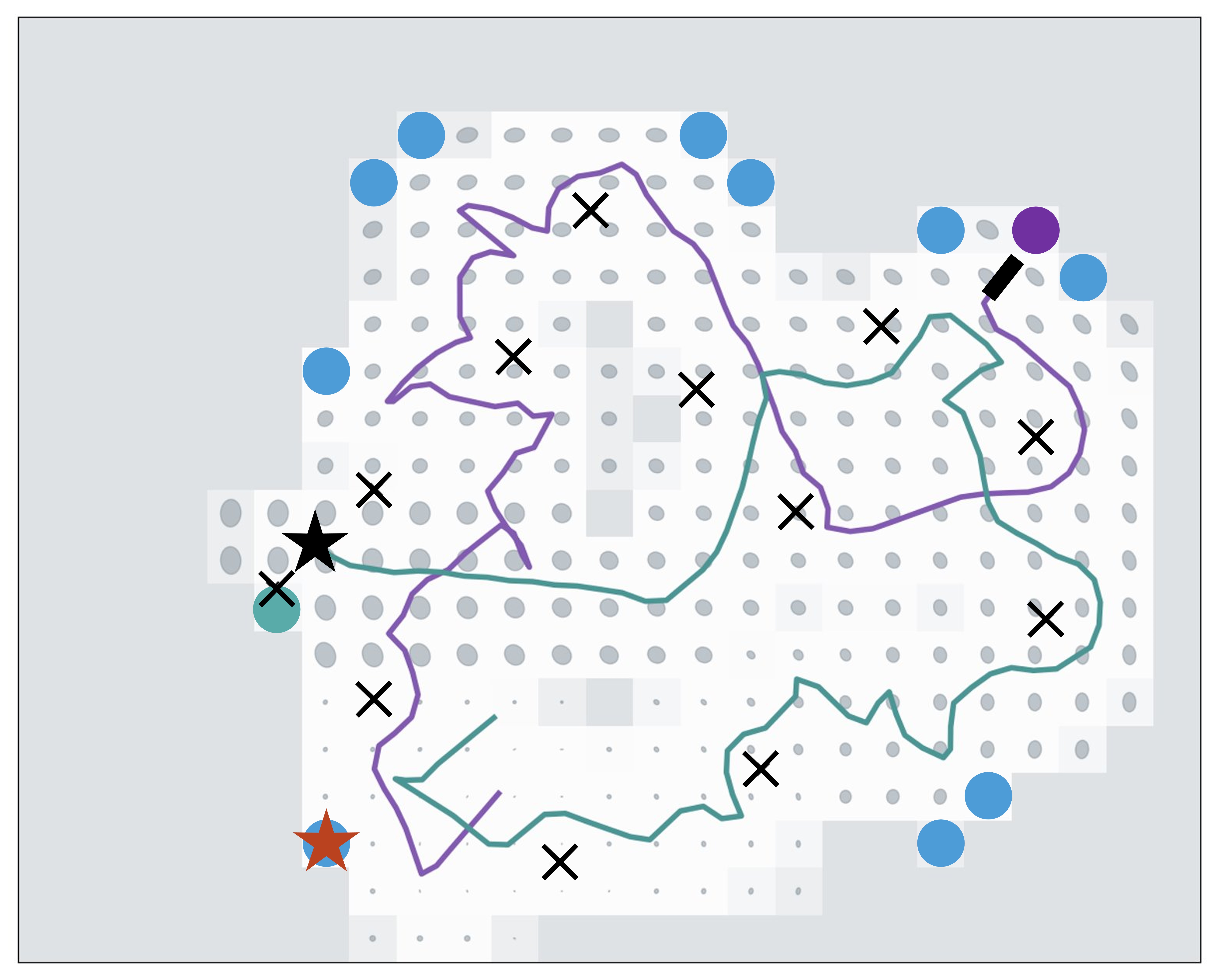}%
    }\hspace{0.2cm}
    \subfigure[Choosing an exploring frontier (green, with red star) due to relatively low map uncertainty.]{%
    \label{fig:obs_a}%
    \includegraphics[width=0.45\columnwidth]{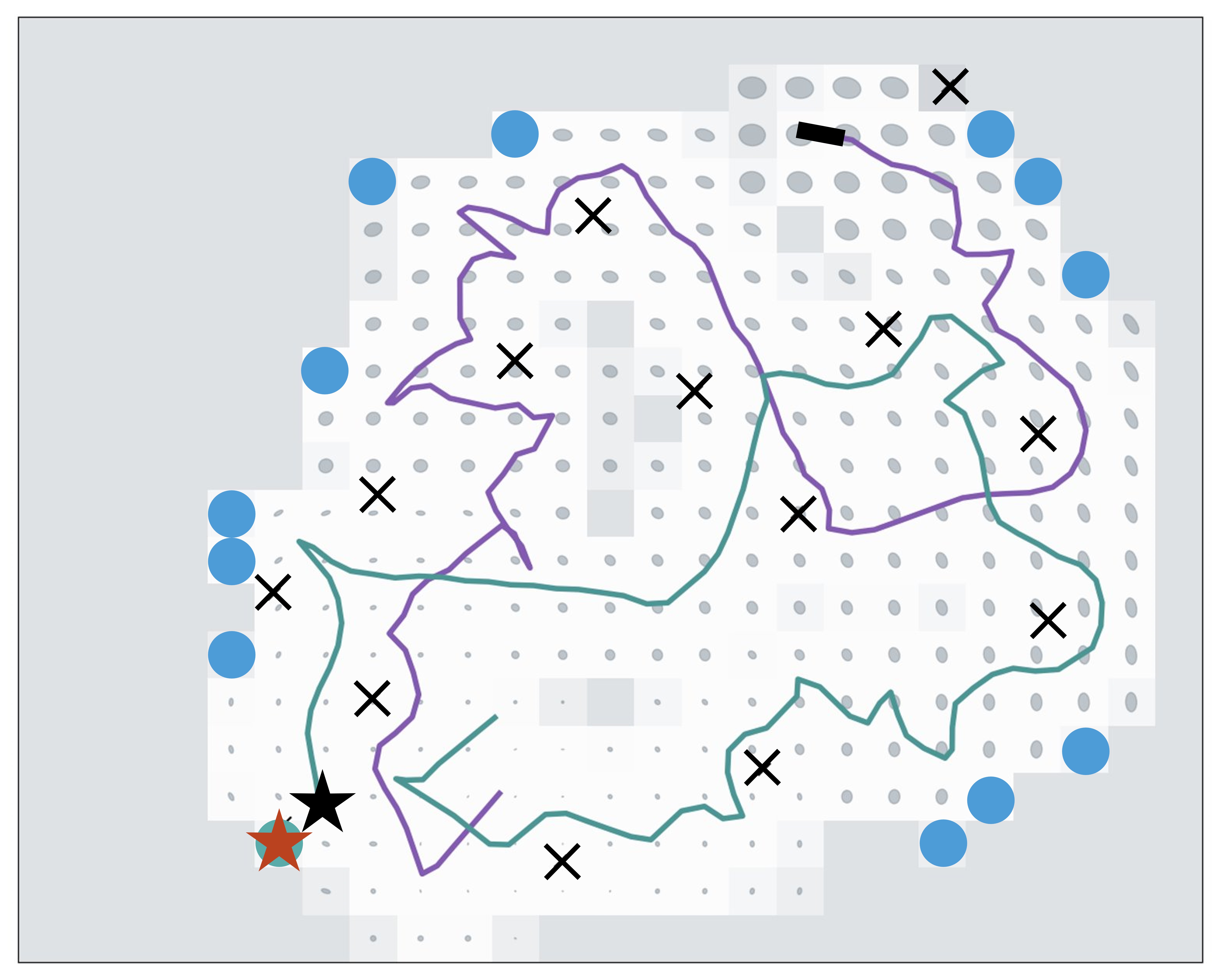}%
  } \vspace{-2mm}
  \caption{\textbf{Problem setup.} In this $100 m \times 80 m$ virtual map created by a two-robot team, gray ellipses depict the uncertainty of visited cells. The green robot's position is denoted by a black star, and its newly selected target state is represented by a red star. The current position of the other robot on the team is marked as a black rectangle. Landmarks are expressed by black x's. Potential frontiers emerge along the boundary between the explored  and unexplored areas. The three types of frontiers—exploring, revisiting, and rendezvous—are denoted by their respective colors: green, blue, and purple.}
\label{fig:virtual-map}%
  \vspace{-6mm}
\end{figure}
Autonomous exploration and mapping have been a vigorously discussed subject for several decades.
Early research has primarily centered around optimizing task distribution among team members \cite{burgard2005coordinated, fox2006distributed}.
As Simultaneous Localization and Mapping (SLAM) techniques have advanced, various approaches \cite{jang2020multi, yu2021smmr, atanasov2015decentralized, kontitsis2013multi, corah2019communication} have incorporated the concept of map uncertainty into autonomous exploration.
These strategies consider Gaussian noise sensor models and Gaussian noise kinematic models, aiming to strike a balance between exploration efficiency and managing uncertainties in the resulting maps.
Although these techniques may differ in terms of their map representation and merging techniques, they share a similar utility function that takes into account both the information gain of the latest generated map and the required travel distance to the waypoints under consideration.
However, in the process of computing information gain, these techniques either ignore uncertainty propagation or have only propagated uncertainty considering future steps, neglecting the history of past steps, which is inadequate for situations with high localization uncertainty.

Building upon our previous work on single robot expectation-maximization (EM) inspired exploration \cite{wang2020autonomous}, we introduce an asynchronous EM exploration algorithm for both centralized and decentralized multi-robot teams. 
The algorithm is tightly coupled with a factor graph SLAM system. 
Upon a robot's arrival at the target position while the exploration remains ongoing, a \textit{virtual map} is constructed based on the SLAM result to gauge the prevailing map uncertainty.  Subsequently, for each potential new target, we execute an expectation-maximization procedure to assess the potential information gain associated with the forthcoming actions of the robot and its interactions with neighboring robots. Selection of a new target goal position is then determined by considering both the collective information gain of the entire team within the virtual map, and the efficiency of task allocation among robots.

Our research introduces several innovative contributions:
\begin{itemize}
\item An asynchronous multi-robot exploration framework catering to both centralized and decentralized SLAM systems, taking into account efficient task allocation for exploration and addressing map uncertainty.
\item Incorporating an expectation-maximization inspired technique to assess the future impact and interactions of a robot with its neighboring entities.
\item Introducing an efficient inter-robot and local map uncertainty propagation approach, 
tailored to scenarios involving multiple robots and localization uncertainty.
\end{itemize}
Our code for the proposed framework is publicly released\footnote{\url{https://github.com/RobustFieldAutonomyLab/Multi-Robot-EM-Exploration}}.
The subsequent sections of our paper are structured as follows: a review of background literature is presented in Sec. \ref{sec:litreview}, followed by the introduction of factor graph SLAM and the proposed expectation-maximization exploration method in Sec. \ref{sec:Formulation}. The comprehensive process of our multi-robot exploration with expectation-maximization is outlined in Sec. \ref{sec:Algorithms}. Our experimental outcomes are detailed in Sec. \ref{sec:experiments}, and the paper concludes with a summary in Sec. \ref{sec:conclusions}.
\section{Related Works}\label{sec:litreview}
In multi-robot exploration, the communication arrangement of the robot system holds significance. 
Charrow et al. \cite{charrow2014approximate} examines the exploration of a small area using a team of robots equipped with range-only sensors using a coordinated approach.
Although a centralized system maximizes the utilization of data collected by robot teams, it experiences increased computation challenges as the size of the robot team expands.
Numerous algorithms, such as \cite{burgard2005coordinated}, \cite{atanasov2015decentralized}, \cite{colares2016next}, \cite{regev2016multi},  \cite{schlotfeldt2018anytime} and \cite{chen2022multi} adopt a decentralized approach. In this approach, robots exchange a significant portion of their historical data while independently making decisions based on their most recent knowledge of the environment.
The asynchronous nature of the decentralized system reduces the communication and computational load in comparison to a scenario involving a centralized computer. However, it can still face limitations imposed by the communication range of the robot team.
Additionally, there exist distributed approaches, discussed in  \cite{fox2006distributed}, \cite{lauri2017multi}, \cite{freda20193d},  \cite{kantaros2019asymptotically} and \cite{ginting2021chord}, in which robots communicate only when they encounter each other, adhering to constraints on communication bandwidth. Their decisions depend on the limited information available from both the robot team and the environment.
Such distributed approaches are designed  for scenarios involving larger robot teams with constrained communication bandwidth, although they may involve some performance trade-offs to accommodate these conditions.

Another critical factor to consider is the criteria utilized for action selection.
In earlier approaches such as  \cite{burgard2005coordinated}, \cite{fox2006distributed}, the primary emphasis was on efficiently distributing tasks among team members.
Later, the ``Next Frontier" \cite{colares2016next} introduced the concept of information potential to improve exploration efficiency.
Jang et al. \cite{jang2020multi} utilize a Gaussian processes to characterize the environment, enabling obstacle avoidance and task allocation for a team of robots.
In NeuralCoMapping \cite{ye2022multi}, a multiplex graph neural network (GNN) is introduced to strike a balance between long-term and short-term performance optimization.
Tzes et al. \cite{tzes2023graph} present a novel approach that integrates the exploration under uncertainty problem into a learning framework using graph neural networks.

Given the uncertainty inherent in both robot control and sensing, some authors incorporate uncertainty into the exploration problem. 
Kontitsis et al. \cite{kontitsis2013multi} incorporate Relative Entropy into the utility function when dealing with a partially known map to reduce localization uncertainty. 
Similarly, in \cite{ossenkopf2019long}, the authors perform joint entropy minimization to actively explore over a long-term horizon. 
Other researchers utilize filter-based techniques to address the uncertainty associated with future steps. In the work by Atanasov et al. \cite{atanasov2015decentralized}, a square-root information filter is employed to predict the impact of future actions.
Schlotfeldt et al. \cite{schlotfeldt2018anytime} adopt a similar filter-based approach within a decentralized system. 
Meanwhile, Kantaros et al. \cite{kantaros2019asymptotically} present a sampling-based method aimed at reducing cumulative uncertainty stemming from dynamic hidden states.
Indelman 
 \cite{indelman2018cooperative} presents an approach that employs belief propagation to anticipate the performance of robot teams in future actions.
Chen \cite{chen2020broadcast} considers graph topology when making action selections.

In contrast to other multi-robot exploration approaches, which may overlook localization uncertainty or prioritize the influence of future robot interactions to reduce map uncertainty, our expectation-maximization based approach emphasizes both the interactions within a robot team and the iterative re-visitation of the existing map. 
Furthermore, we use a \textit{virtual map}, which serves as a robust tool for assessing not only the localization uncertainty of individual robots but also the collective uncertainty introduced by robots into the map.
\section{Problem Formulation and Approach}\label{sec:Formulation} 
We address an autonomous exploration problem that is tightly coupled with a SLAM factor graph for a team of $n$ robots. 
We make the assumption that the initial states of all robots are sufficiently close to each other, enabling mutual observation among group members and facilitating an efficient map initialization process.
Additionally, we impose a boundary on the exploration task, where the exploration process terminates upon fully exploring the enclosed environment.
\subsection{Simultaneous Localization and Mapping}
Let ${N} = \{1, 2, \cdots, n\}$ be the set of $n$ robots. 
For each robot $\alpha \in N$, we denote its state at timestamp $i$ as $\mathbf{x}_{\alpha, i}$. 
The robot odometry observation between present state $\mathbf{x}_{\alpha, i}$ and previous state $\mathbf{x}_{\alpha, i-1}$ is described by the equation:
\begin{align}\label{sensor1}
    \mathbf{z}^{\alpha,i-1}_{\alpha, i} =f(\mathbf{x}_{\alpha, i-1}, \mathbf{x}_{\alpha, i}) + \epsilon^{\alpha,i-1}_{\alpha, i}.
\end{align}
Assume robot $\alpha$ observes a landmark state $\mathbf{l}_j$ at timestamp $i$, we can describe the landmark observation:
\begin{align}\label{sensor2}
    \mathbf{z}^{\alpha, i}_{j} =g(\mathbf{x}_{\alpha, i}, \mathbf{l}_j) + \epsilon^{\alpha, i}_{j}.
\end{align}
If another robot $\beta$ is observed at timestamp $i$ by robot $\alpha$,we refer to this as a \textit{robot rendezvous observation}:
\begin{align}\label{sensor3}
    \mathbf{z}^{\alpha, i}_{\beta, {i}} =f(\mathbf{x}_{\alpha, i}, \mathbf{x}_{\beta, {i}}) + \epsilon^{\alpha, i}_{\beta, {i}},
\end{align}
where $f(\cdot)$ denotes the state transformation between robot states, $g(\cdot)$ is the state transformation from robot state to landmark state, and $\epsilon^{\alpha,i-1}_{\alpha, i}$, $\epsilon^{\alpha, i}_{j}$ and $\epsilon^{\alpha, i}_{\beta, {i}}$ are zero-mean Gaussian noise variables.

At present timestamp $t$, $\mathcal{X} = \{\mathbf{x}^{\alpha}_i | \alpha \in {N}, i \in [0,t]\}$ represents the set containing the states of all $n$ robots from the initial timestamp $0$ to the present timestamp $t$. $\mathcal{L} = \{\mathbf{l}_0, \mathbf{l}_1, \ldots, \mathbf{l}_m\}$ is the set of landmarks observed until timestamp $t$. 
In this paper we define a landmark as a point with a unique identifier that is observable by the robots' on-board sensors.
Additionally, let $\mathcal{Z}$ be the set containing odometry observations, landmark observations and robot rendezvous observations from all timestamps.
The SLAM problem can be framed as a maximum a posteriori estimation problem \cite{kaess2012isam2}:
\begin{align}\label{slam_eqn}
    \mathcal{X}^*,\mathcal{L}^* &= \mathop{\arg\max}_{\mathcal{X},\mathcal{L}} P(\mathcal{X}, \mathcal{L}| \mathcal{Z}).
\end{align}
\subsection{Expectation-Maximization Exploration} \vspace{-2mm}
During the exploration process, when a robot $\alpha$ reaches its current target state $\mathbf{a}_{\alpha}^{\text{this}}$, it becomes necessary to select a new target state $\mathbf{a}_{\alpha}^{\text{next}}$ from a set of potential new states $\mathcal A_{\alpha}^{\text{next}}$. 
Building upon our previous research on single-robot exploration \cite{wang2020autonomous}, we consider a frontier-based strategy that incorporates two key factors: efficient task allocation among robots \cite{burgard2005coordinated} and the maintenance of a low-uncertainty map.

When $\alpha$ reaches target state $\mathbf{a}^{\text {this}}_{\alpha}$, $\forall {\mathbf{a}^{\text{next}}_{\alpha}}' \in \mathcal A_{\alpha}^{\text{next}}$, we define:
\begin{align} \vspace{-3mm}
\mathcal{X^{\text{new}}} = \mathcal{X}^{\text{old}} \cup \mathcal{X}^{\text{predict}} \cup \mathcal{X}^{\text{next}}.
\vspace{-1mm} \end{align} 
$\mathcal{X}^{\text{old}}$ contains the historical states of all $n$ robots, $\mathcal{X}^{\text{predict}}$ denotes the set of predicted robot states for each current target state $\{\mathbf{a}^{\text{this}}_i | i \in N, i \neq \alpha \}$ and $\mathcal{X}^{\text{next}}$ represents the state sequence of robot $\alpha$ resulting from ${\mathbf{a}^{\text{next}}_{\alpha}}'$.
A classification EM algorithm is used to predict the change of map uncertainty due to ${\mathbf{a}^{\text{next}}_{\alpha}}'$. 
To avoid the exponential expansion in potential virtual landmark states, we substitute the \textbf{E-step} with a classification step (\textbf{C-step}), in which we construct a virtual map using the historical data of our robot team: 
\begin{align} \vspace{-1mm}
\mathcal{V}^* &= \mathop{\arg\max}_{\mathcal{V}} P(\mathcal{V}| \mathcal{X}^{\text {old}}, \mathcal{Z}^{\text{old}}),\\\label{eqn:e-step}
&={M}({\mathcal{X}^{\text {old}}}^{*}, \mathcal{Z}^{\text{old}}).
\vspace{-1mm} \end{align} 
Here, $\mathcal{Z}^{\text{old}}$ represents the observations associated with $\mathcal{X}^{\text{old}}$.
${\mathcal{X}^{\text {old}}}^{*}$ is the optimized value from the previous SLAM optimization.
${M}(\cdot)$ is the inverse observation model used to estimate the mean and covariance of the virtual map with the given SLAM estimate ${\mathcal{X}^{\text {old}}}^{*}$.
For a full explanation of the virtual map construction process, please refer to our prior work \cite{wang2022virtual}. Subsequently, in the \textbf{M-step}, we again perform maximum a posteriori estimation to find $\mathcal{X}^{\text{new}}$ (and choose $\mathcal{X}^{\text{next}})$:
\vspace{-2mm}
\begin{align}
{\mathcal{X}^{\text{new}}}^* = \mathop{\arg\max}_{\mathcal{X}^{\text{new}}} P(\mathcal{X}^{\text{new}}| \mathcal{V}^*, \mathcal{Z}^{\text{new}}) ,\\
\mathcal{Z}^{\text{new}} = \mathcal{Z}^{\text{old}} \cup \mathcal{Z}^{\text{predict}}.
\end{align}
The updated observation set, $\mathcal{Z}^{\text{new}}$, is a combination of both previously gathered observations, $\mathcal{Z}^{\text{old}}$, and anticipated future observations, $\mathcal{Z}^{\text{predict}}$. 
These predicted observations are determined via the measurement models outlined in Eq. (\ref{sensor1}), Eq. (\ref{sensor2}), and Eq. (\ref{sensor3}), based on the virtual map $\mathcal{V}^*$ and the target state assigned to each robot.
Thus, we assess the overall local map uncertainty of robot $\alpha$ by computing the sum of the individual uncertainties for each map element in the virtual map $\mathcal{V} (\cdot)$ derived from Eq. (\ref{eqn:e-step}), considering only the optimized states of robot $\alpha$, ${\mathcal{X}^\text{new}_{\alpha}}^*$:
\begin{align}\label{U_M}
U_M &= \phi (\Sigma_{{M}({\mathcal{X}^\text{new}_{\alpha}}^*, \mathcal{Z}^{\text{new}})}),\\
 &= \sum_{\mathbf{v_i} \in  {M} ({\mathcal{X}^\text{new}_{\alpha}}^*, \mathcal{Z}^{\text{new}})} \phi (\Sigma_{\mathbf{v}_i}).
\end{align}
Each element $\mathbf{v}_i$ contributes to the sum based on its covariance $\Sigma_{v_i}$, and in this paper, we employ the A-Optimality metric as our uncertainty criterion $\phi$.

To optimize the distribution of exploration tasks among the robots, we utilize a distance evaluation metric inspired by the methodology presented in \cite{burgard2005coordinated}. 
For each potential new target state ${\mathbf{a}_{\alpha}^t}'$ assigned to robot $\alpha$ at time $t$, we quantify the effectiveness of task allocation using $U_T$.
A higher $U_T$ results in the robots being closer to their teammates, increasing the likelihood that they repeatedly explore the same area.
\begin{align}
    U_T = \sum_{
    \mathbf{a}_{\beta}^i \in \mathcal{A}, \beta \neq \alpha
    } 
    h(\| {\mathbf{a}_{\alpha}^t}^{'} - \mathbf{a}_{\beta}^i \|_2), 
     \\
    h(d) = \begin{cases}
    1 - \frac{d}{d_{\text{max}}} & d < d_{\text{max}} \\
    0 & d \geq d_{\text{max}}
    \end{cases}.
\end{align}
Here, $\left\| \cdot \right\|_2$ represents the L2 norm, and $\mathcal{A}$ denotes the collection of all historical target states for all robots. Based on this, we can define the new target state for robot $\alpha$ as one of the potential target states that optimally balances the factors $U_M$, $U_T$ and the Euclidean distance to target state factor, $U_D$, with scale factors $\lambda_0$, $\lambda_1$, $\lambda_2$:
\begin{align}\label{utility}
    \mathbf{a}^{\text{next}}_{\alpha} = \mathop{\arg\min}_{{\mathbf{a}_{\alpha}^t}'}(\lambda_0 U_M + \lambda_1 U_T + \lambda_2 U_D).
\end{align}

\section{Proposed Multi-Robot Exploration Algorithm}\label{sec:Algorithms}
\begin{figure}[t]
  \centering
    \includegraphics[width=.75\columnwidth]{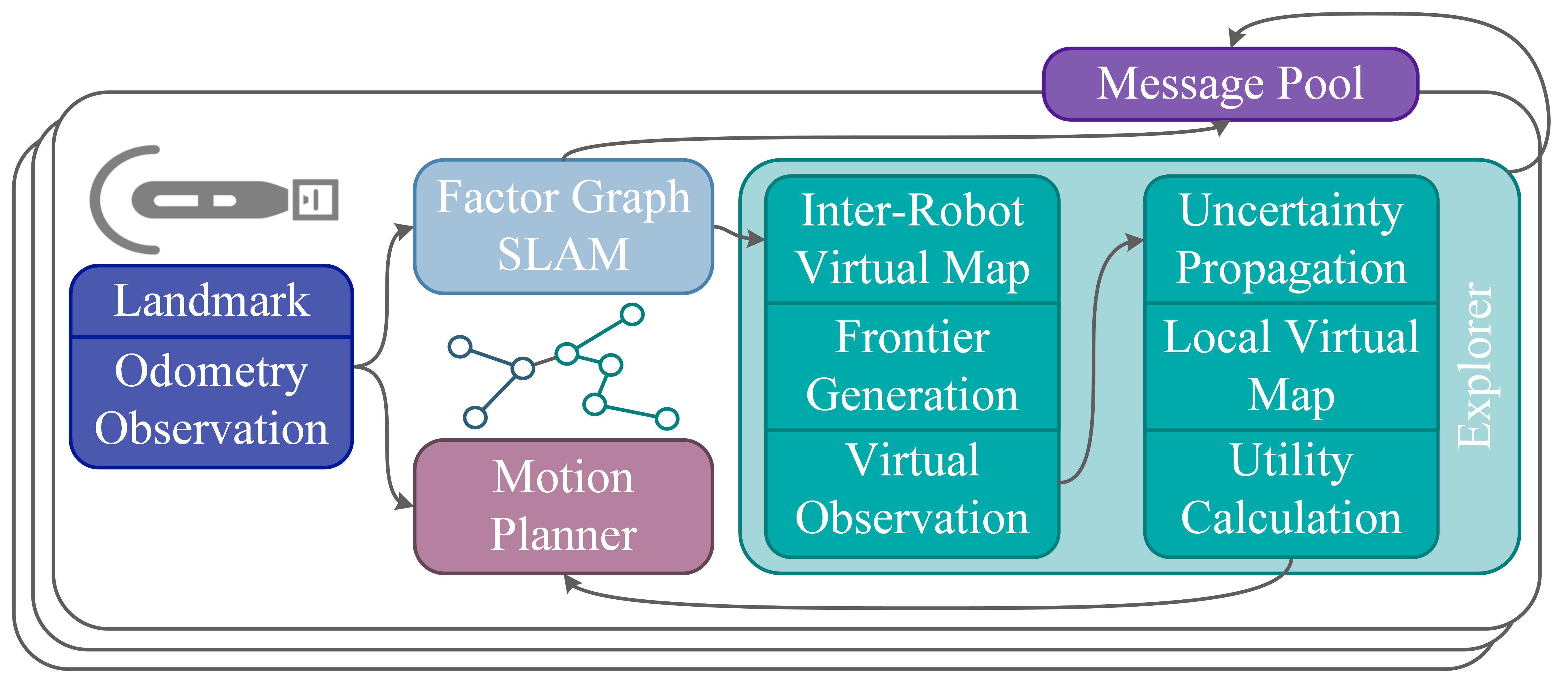}%
    \vspace{-2mm}
  \caption{\textbf{System Architecture.} The pipeline of the proposed approach.}
\label{fig:pipeline}%
\vspace{-6mm}
\end{figure}

Here we discuss the details of the proposed 
algorithm.
Fig. \ref{fig:pipeline} shows the pipeline of the proposed algorithm.
We consider a situation where a group of robots collaborates, and each individual robot within the group is furnished with both motion and perception sensors.
A centralized or decentralized factor-graph based SLAM algorithm is assumed to operate at a specific frequency. 
The robot team shares their optimized SLAM trajectories, landmark positions and history of chosen target states among its members. 
This simulation does not take into account any limitations on communication bandwidth.
However, individual robots maintain virtual maps locally using the latest SLAM estimates.

After a robot $\alpha$ successfully reaches its current target state, its virtual map is updated and a group of potential new target states $\mathcal A_{\alpha}^{\text{next}}$ is chosen from the virtual map.
Subsequently, the virtual observation is synthesized utilizing the prevailing environmental knowledge.
Next, the expectation-maximization based uncertainty propagation computes the potential impact of the target state under consideration, leading to an update in the virtual map's covariances. 
The new target goal is then selected according to the utility function (Eq. (\ref{utility})).
Finally, the motion planner is initiated to formulate a series of actions leading robot $\alpha$ to the new target state.

\vspace{-1mm}

\subsection{Virtual Map}
As depicted in Eq. (\ref{eqn:e-step}), the virtual map $\mathcal V$ of a finite environment is generated from the robot states $\mathcal{X}$ and their associated observations $\mathcal Z$. 
Assuming that $\mathcal V$ comprises $b$ map cells $\mathbf{v}_i$ referred to as \textit{virtual landmarks}, the posterior can be redefined as follows:
\begin{align}
P(\mathcal{V}| \mathcal{X}, \mathcal{Z}) = \prod_{\mathbf{v}_i\in \mathcal{V}} P (\mathbf{v}_i| \mathcal{X}, \mathcal{Z}).
\end{align}
The likelihood of virtual landmark ${\mathbf{v}_i}$ being observed is  $q({\mathbf{v}_i}) = \mathbb E[P(\mathbf{v}_i|\mathcal{X, Z})]$. 
When this virtual landmark is observed by multiple robot states, the procedure for updating $q({\mathbf{v}_i})$ is similar to updating map cell values in an occupancy grid map \cite{thrun2002probabilistic}.
We assume the virtual landmark $\mathbf{v}_i$ is observed by a robot state $\mathbf{x}_{\alpha, j}$, with its estimated value denoted $\hat{\mathbf{x}}_{\alpha, j}$, and a marginal covariance $\Sigma_{\mathbf{x}_{\alpha, j}}$. We can compute the covariance: $
\Sigma_{\mathbf{v}_i} = \mathbf{H} \cdot \Sigma{\mathbf{x}_{\alpha, j}} \cdot \mathbf{H}^\intercal$. 
Here, $\mathbf{H} = \frac{\partial g(\mathbf{x}_{\alpha, j}, \mathbf{v}_i)}{\partial \mathbf{x}_{\alpha, j}} |  \hat{\mathbf{x}}_{\alpha, j}$ represents the Jacobian matrix obtained by differentiating the landmark observation model $g(\mathbf{x}_{\alpha, j}, \mathbf{v}_i)$ with respect to estimated robot state $\hat{\mathbf{x}}_{\alpha, j}$.
We utilize Covariance Intersection to compute the covariance of a virtual landmark that is observed by multiple robot states, as detailed in \cite{wang2022virtual}.
Fig. \ref{fig:virtual-map} shows inter-robot virtual maps built from both local robot states and neighbors' robot states received by local robot $\alpha$. The observed regions are highlighted in white; gray ellipses show covariances describing the uncertainty of the map's cells.

\vspace{-1mm}

\subsection{Uncertainty Propagation}
\begin{figure}
\centering
\includegraphics[width=0.8\columnwidth]{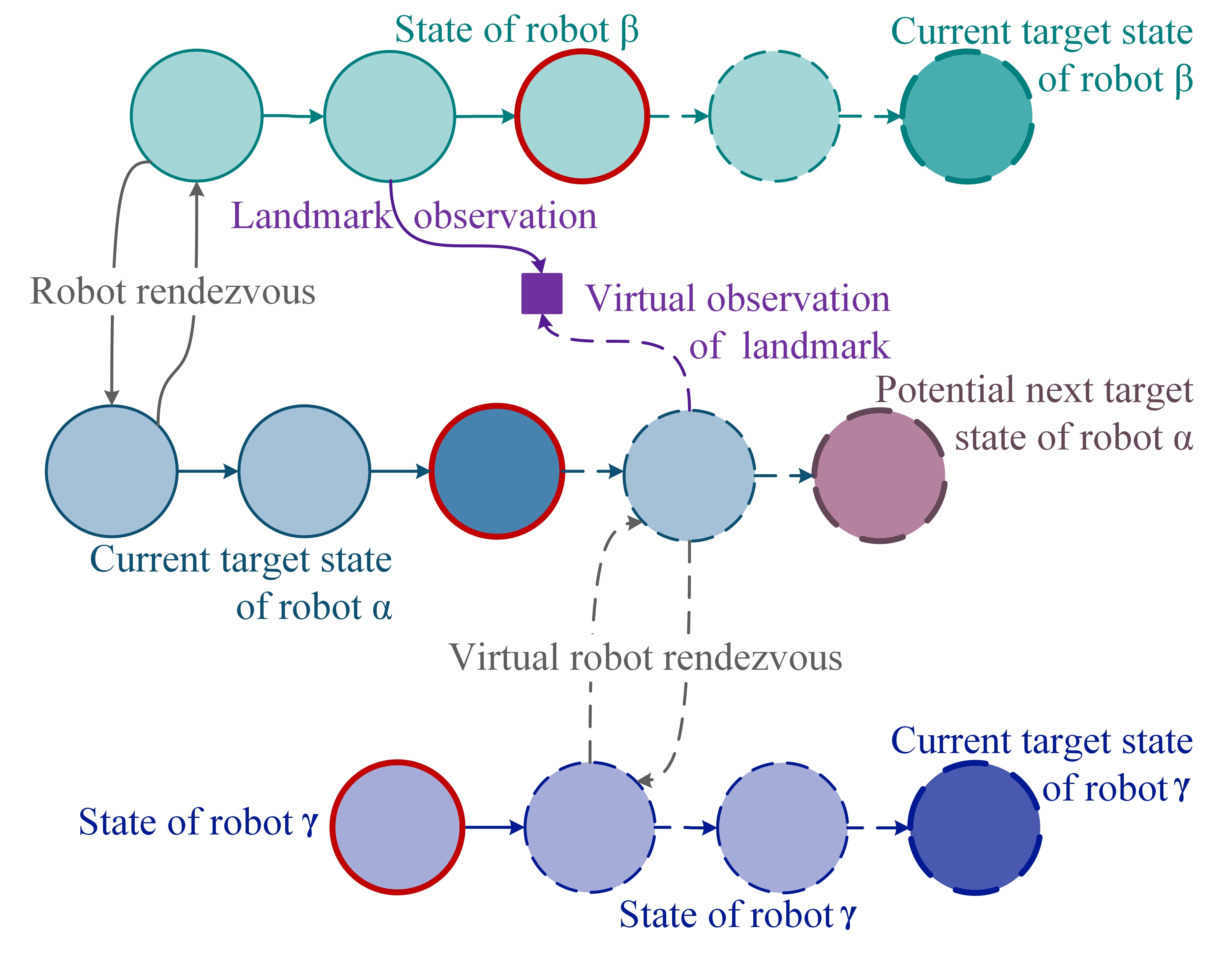}%
\vspace{-2mm}
  \caption{\textbf{EM-based uncertainty propagation with virtual observations.} Nodes representing the current robot states are distinguished by red edges. For every potential next target state (shown in pink), a trajectory simulation is executed, leading to the generation of a sequence of virtual observations indicated by dashed arrows. Simultaneously, we model the future states and observations of other robots as they approach their individual current target states.
  }\label{fig:uncertainty} 
  \vspace{-6mm}
\end{figure}
\begin{algorithm}[t]
\SetAlgoNoEnd
\SetKwFunction{GVW}{GenerateVirtualWaypoints}{}
\SetKwFunction{VO}{VirtualObserve}{}
\SetKwFunction{UG}{UpdateGraph}{}
\SetKwFunction{OG}{OptimizeGraph}{}
\textbf{Global:} Latest SLAM Graph $\mathcal{G}$,andInter-robot Virtual Map $\mathcal{V}^*$ \\
\KwIn{Potential frontier state ${\mathbf{a}_{\alpha}^t}'$, Current robot states $\mathcal{X}^t$, Current Target States of Neighbors $\mathcal A^t$}
\KwOut{Optimized robot states $\mathcal{{X^{\text{new}}}^*}$}
$\mathcal{X}^{\text{predict}} \gets \emptyset $\\
\ForEach{$\mathbf{a}_{\gamma}^t \in \mathcal{A}^t, 
\mathbf{x}_{\gamma}^t \in \mathcal{X}^t$}{
$\mathcal{X}_{\gamma}^{t:\ast}$ $\gets$
    \GVW{$\mathbf{a}_{\gamma}^t, \mathbf{x}_{\gamma}^t$}\\
    $\mathcal{X}^{\text{predict}} \gets \mathbf{X}_{\gamma}^t \cup \mathcal{X}^{\text{predict}} $
}
$\mathcal X^{\text{next}} \gets $\GVW{${\mathbf{a}_{\alpha}^t}', \mathbf{x}_{\gamma}^t$}\\
{\color{gray} \# Calculate virtual observations}\\
$\mathcal{Z}^{\text{predict}} \gets$ \VO{$\mathcal{V}^*, \mathcal{X}^{\text{predict}} \cup \mathcal{X}^{\text{next}}$}\\
{\color{gray} \# M-step: graph optimization}\\
\UG{$\mathcal{G}, \mathcal{X}^{\text{predict}} \cup \mathcal{X}^{\text{next}}$}\\
$\mathcal{{X^{\text{new}}}^*}$ $\gets$ \OG{$\mathcal{G}$}\\
\KwRet{${\mathcal{X}^{\mathrm{new}}}^*$}
\caption{Uncertainty Propagation}\label{alg1}
\vspace{-7mm}
\end{algorithm}
As depicted in Fig. \ref{fig:virtual-map}, most potential target states are chosen from the perimeters of the observed regions. Three types of frontiers are identified for selection: \textit{exploration frontiers} close to the robot's latest position, \textit{revisiting frontiers} near previously visited landmarks, and \textit{rendezvous frontiers}, which are the current target positions of neighboring robots.
Subsequently, for each potential target, a set of waypoints $\mathcal{X}^{\text{new}}$ is uniformly sampled along the shortest path connecting the robot's current state and the next potential target state. Additionally, waypoints $\mathcal{X}^{\text{predict}}$ are generated to connect the present states of all robots with their respective target states. 
The process of generating virtual observations $\mathcal{Z}^{\text{predict}}$ along the paths to these target states is depicted in Fig. \ref{fig:uncertainty}. Virtual odometry measurements are created between adjacent waypoints using Eq. (\ref{sensor1}). For virtual landmark observations, Eq. (\ref{sensor2}) is employed, generating observations between previously observed landmarks and nearby waypoints. When two robots are within each other's sensing range at the same timestep, a virtual robot observation is produced using Eq. (\ref{sensor3}).
Following this, virtual observations are added into the SLAM graph to propagate uncertainty. The procedure is outlined in Alg. \ref{alg1}.

\vspace{-2mm}

\subsection{Map Uncertainty Utility Computation} \vspace{-1mm}
We compute the uncertainty of map cells by considering both the likelihood of the existing virtual map $\mathcal{V}^*$ and the optimized robot states of robot $\alpha$, ${\mathcal{X}^{\text{new}}_{\alpha}}^* \subset {\mathcal{X}^{\text{new}}}^*$. Referencing Fig. \ref{fig:obs_a}, it is evident that despite the purple robot's trajectory being impacted by localization uncertainty due to accumulated odometry error, this uncertainty is partially mitigated by the historical trajectory of the green robot, which exhibits relatively lower uncertainty. The construction of an inter-robot virtual map for map uncertainty estimation could potentially lead to conflicts in decision-making for a local robot.
As outlined in Alg. \ref{alg2}, we generate a local virtual map denoted as ${\mathcal{V}^{\mathrm{new}}_\alpha}^*$ for robot $\alpha$. This map is constructed using ${\mathcal{X}^{\text{new}}_{\alpha}}^*$ and corresponding virtual observations ${\mathcal{Z}^{\text{new}}_{\alpha}}$. Then, we compute the uncertainty utility of the map by considering the overlapping observed regions common to both the current inter-robot virtual map ${\mathcal V}^*$ and the predicted local virtual map ${\mathcal{V}^{\mathrm{new}}_\alpha}^*$, via Eq. (\ref{U_M}).

\vspace{-2mm}

\subsection{Complexity Analysis}
In this section, we analyze the time complexity of the Expectation-Maximization Explorer shown in Fig. \ref{fig:pipeline}.
For a team of $n$ robots, each with a maximum of $N_x$ historical robot states, updating a robot state in the virtual map takes $T_v$ time, resulting in a potential inter-robot virtual map construction time of $n\cdot N_x\cdot T_v$.
During frontier generation, detecting exploration bounds (time $T_b$) containing $N_c$ virtual landmarks and observing each landmark (time $T_l$) contributes to a step duration of $T_b + N_c\cdot T_l$.
Frontier selection generates $N_f$ frontiers. Crafting virtual waypoints (up to $N_w$) and conducting virtual robot rendezvous checks (time $T_r$) per frontier results in a total frontier generation time of $(N_f + n - 1)\cdot(T_l + T_r)$.
Utilizing iSAM2 \cite{kaess2012isam2} for uncertainty propagation (time $T_u$  per update) leads to a combined propagation time of $N_f \cdot T_u$.
Constructing a local virtual map requires $N_x \cdot T_v$ time.
The most time-intensive steps involve iSAM2's uncertainty propagation ($\mathcal{O}(n^{2.36})$ complexity for $n$ states) and covariance intersection during virtual map creation ($\mathcal{O}(s^3)$ complexity with $s$ nonzero matrix block size) \cite{kaess2009covariance}.

\vspace{-3mm}
\begin{algorithm}[t]
\SetAlgoNoEnd
\SetKwFunction{trace}{trace}
\SetKwFunction{q}{q}
\KwIn{
Inter-robot Virtual Map $\mathcal{V}^*$,
Local Virtual Map ${\mathcal{V}^{\mathrm{new}}_\alpha}^*$}
\KwOut{Map uncertainty utility factor $U_M$}
$U_M \gets 0$\\
\For{$i \leftarrow 0$ \KwTo $b$}{
    $\mathbf{v}_{i,g} \in \mathcal{V}^* $, 
    $\mathbf{v}_{i,l} \in {\mathcal{V}^{\mathrm{new}}_\alpha}^* $\\
    {\color{gray} \# $q_{\mathrm {min}}$: minimum accepted probability of observed.} \\
    \If{ \q{$\mathbf{v}_{i,l}$} $> q_{\mathrm{min}}$ \textbf{and} \q{$\mathbf{v}_{i,g}$} $> q_{\mathrm{min}}$}{
        $U_M \gets U_M$ + \trace{$\Sigma_{\mathbf{v}_{i,l}}$} \color{gray} \color{gray}{\# A-Optimality}
    }
}
\KwRet{$U_M$}
\caption{Compute $U_M$ of potential frontier ${\mathbf{a}_{\alpha}^t}'$}
\label{alg2}
\vspace{-9mm}
\end{algorithm}
\vspace{0mm}
\section{Experiments and Results}\label{sec:experiments}
\vspace{-1mm}
\begin{figure*}
  \centering
  \includegraphics[width=2\columnwidth]{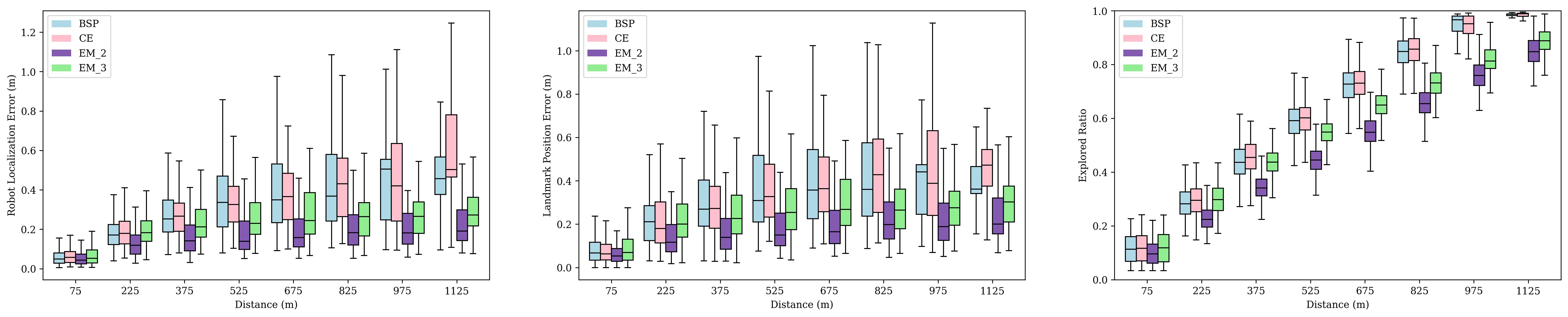}%
  \vspace{-2mm}
  \caption{\textbf{50 trials of three robots navigating in 100m x 100m environments with 20 landmarks, each with a radius of 1m.} They are exploring the environment by constructing a virtual map with cell size of $2 m$. At left, the average robot localization error for each robot state, at center, the average landmark position error for each landmark, and at right, the explored ratio, all plotted against distance.}
\label{fig:small_q}%
\vspace{-4mm}
\end{figure*}
\begin{figure*}
  \centering
  \includegraphics[width=2\columnwidth]{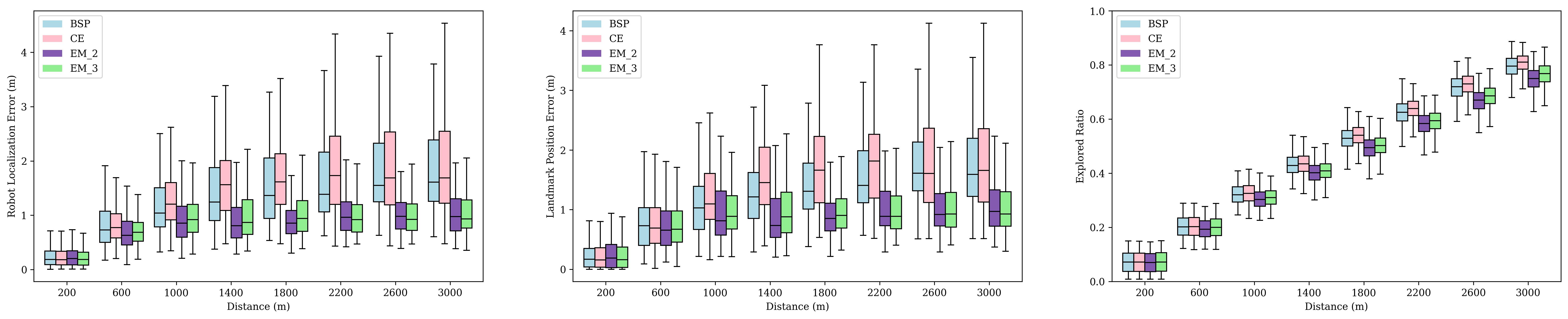}%
  \vspace{-2mm}
  \caption{\textbf{50 trials of three robots navigating in 200m x 200m environments with 20 landmarks, each with a radius of 1m.} They are exploring the environment by constructing a virtual map with cell size of $4 m$. At left, the average robot localization error for each robot state, at center, the average landmark position error for each landmark, and at right, the explored ratio, all plotted against distance.}
\label{fig:big_q}%
\vspace{-6mm}
\end{figure*}
We perform simulated experiments within environments of varied size, employing randomly generated landmarks. The errors described below define 95\% confidence intervals. Similar to our previous work \cite{wang2020autonomous}, we assume each robot is furnished with a sonar with range error 0.002m, bearing error 0.5$\degree$, and max. sensing range 7.5m.
Each robot also performs inertial dead reckoning with  translational and rotational components in a 2D space.
The gyro and accelerometer measurements introduce errors of 0.5 deg. and 0.05 meters, respectively.
The robot has the capability to rotate by $15 \degree$ during each action, maintaining a constant speed of $1 m/s$. 
It is restricted to moving solely in the direction of its current heading.
At the beginning of each trial, we ensure that all robots are located within the sensing range of their teammates to guarantee a proper initialization of the SLAM framework.
To navigate towards uncharted territory, we utilize the Artificial Potential Field (APF) method, detailed in \cite{fan2020improved}, to avoid collision.
The boundaries of the environment are restricted from selection as frontiers, preventing robots from exiting the mission area.

We compare our proposed approach, denoted EM, with two advanced multi-robot exploration algorithms: the coordinated multi-robot exploration method by Burgard et al. \cite{burgard2005coordinated}, referred to as CE in subsequent sections, and the coperative multi-robot belief space planning technique introduced by Indelman et al. \cite{indelman2018cooperative}, denoted as BSP subsequently. Identical frontier selection criteria and virtual observation generation techniques are employed for both methods.

\textbf{The CE planner} \cite{burgard2005coordinated} places emphasis on optimizing task distribution among robots, employing a utility function that takes task allocation into account:
\begin{align}
U_{CE} = \lambda_0 U_D + \lambda_1 U_T.
\end{align}
In this context, both $U_D$ and $U_T$ are computed the same way as outlined in Eq. \ref{utility}.
In our experiment, we select $\lambda_0 = 1$ and $\lambda_1 = 10$.
\textbf{The BSP planner} \cite{indelman2018cooperative} takes into account both exploration efficiency and localization uncertainty, employing a similar virtual observation strategy as ours. This leads to the formulation of the following utility function:
\begin{align} \vspace{-2mm}
U_{BST} &= \lambda_0 U_D + \lambda_1 U_M,\\
U_M &= \sum_{\mathbf{x}_i \in \mathcal{{X^{\text{predict}}}^*}\cup \mathcal{{X^{\text{next}}}^*}} {\phi_A (\sqrt{\Sigma_{\mathbf{x}_i}})}.
\end{align}
We choose $\lambda_0$ as 5 and $\lambda_1$ as 1. Here, $\Phi_{A}(\cdot)$ denotes the trace of the matrix, and ${\mathcal{X}^{\text{predict}}}^*$ and ${\mathcal{X}^{\text{next}}}^*$ are computed through the maximum a posteriori process: \hspace{30mm} (19)
\begin{align*}
\mathcal{X^{\text{predict}}}^*, \mathcal{X^{\text{next}}}^* = \mathop{\arg\max}_
{\mathcal{X^{\text{predict}}},
\mathcal{X^{\text{next}}}} P(\mathcal{X^{\text{predict}}},
\mathcal{X^{\text{next}}}
| \mathcal{Z}^{\text{predict}}, \mathcal{L}). \vspace{-2mm}
\end{align*}
The BSP planner, focusing only on the localization uncertainty of future steps, ignores potential benefits from revisiting previously explored landmarks by the team.
Since only the uncertainties at individual robot states are taken into account, $U_M$ might become imbalanced and not effectively depict the overall localization uncertainty across the entire environment, especially in the presence of varying vehicle speeds.

A quantitative analysis using two distinct sizes of random generalized environments is conducted: one measuring $100m \times 100m$, and the other measuring $200m \times 200m$. 
The $200m \times 200m$ environment presents a unique challenge due to its larger size and sparser distribution of landmarks.
Across 50 trials, each trial involves 20 landmarks with a radius of $1m$ within the environment. 
We introduce three robots into the environment to prevent rendezvous and landmark revisitation from becoming trivially achievable in scenarios involving a large number of robots.
The landmark positions are selected randomly, ensuring a minimum distance of $10m$ between each pair of landmarks.
For each trial, the robots initiate their positions from the middle left region of the environment, with their initial positions also being chosen randomly. 

We test two distinct configurations for the proposed EM explorer: one with $\lambda_0 = 1, \lambda_1 = 0, \lambda_2 = 10$ referred to as EM\_2, and the other with $\lambda_0 = 1, \lambda_1 = 20 \cdot (1 - r), \lambda_2 = 10$ referred to as EM\_3, where $r$ denotes the ratio of the explored area in the environment. For the $100m \times 100m$ environment, we employ a virtual map cell size of $c_v = 2m$, while for the $200m \times 200m$ environment, we choose a cell size of $c_v = 4m$ for the virtual map.
We use three statistics 
to evaluate the performance of the proposed algorithm:
\begin{itemize}
    \item Robot localization error: root-mean-square error (RSME) of the optimized trajectories of all robots from the SLAM framework.
    \item Landmark position error: RSME of the positions of all landmarks observed by the robot teams.
    \item Explored ratio: The proportion of the area considered as observed in the inter-robot virtual map.
\end{itemize}
Fig. \ref{fig:small_q} and Fig. \ref{fig:big_q} depict the outcomes across different scenarios within smaller and larger environments, respectively. EM\_2 and EM\_3 exhibit superior accuracy in localizing both robot and landmark states when compared to CE and BSP. 
This improvement stems from the fact that the proposed approach takes into account not only the forthcoming interactions among robots but also the impact of the team's historical actions.
Nonetheless, it's worth noting that the proposed method demonstrates lower exploration efficiency when contrasted with both CE and BSP, especially in smaller environments. This trade-off is made in favor of achieving higher accuracy.
When considering the task allocation element, the proposed approach in the EM\_3 configuration demonstrates better exploration efficiency compared to the EM\_2 configuration.

\begin{figure}[t]
  \centering
  \subfigure{%
    \includegraphics[width=0.4075\columnwidth]{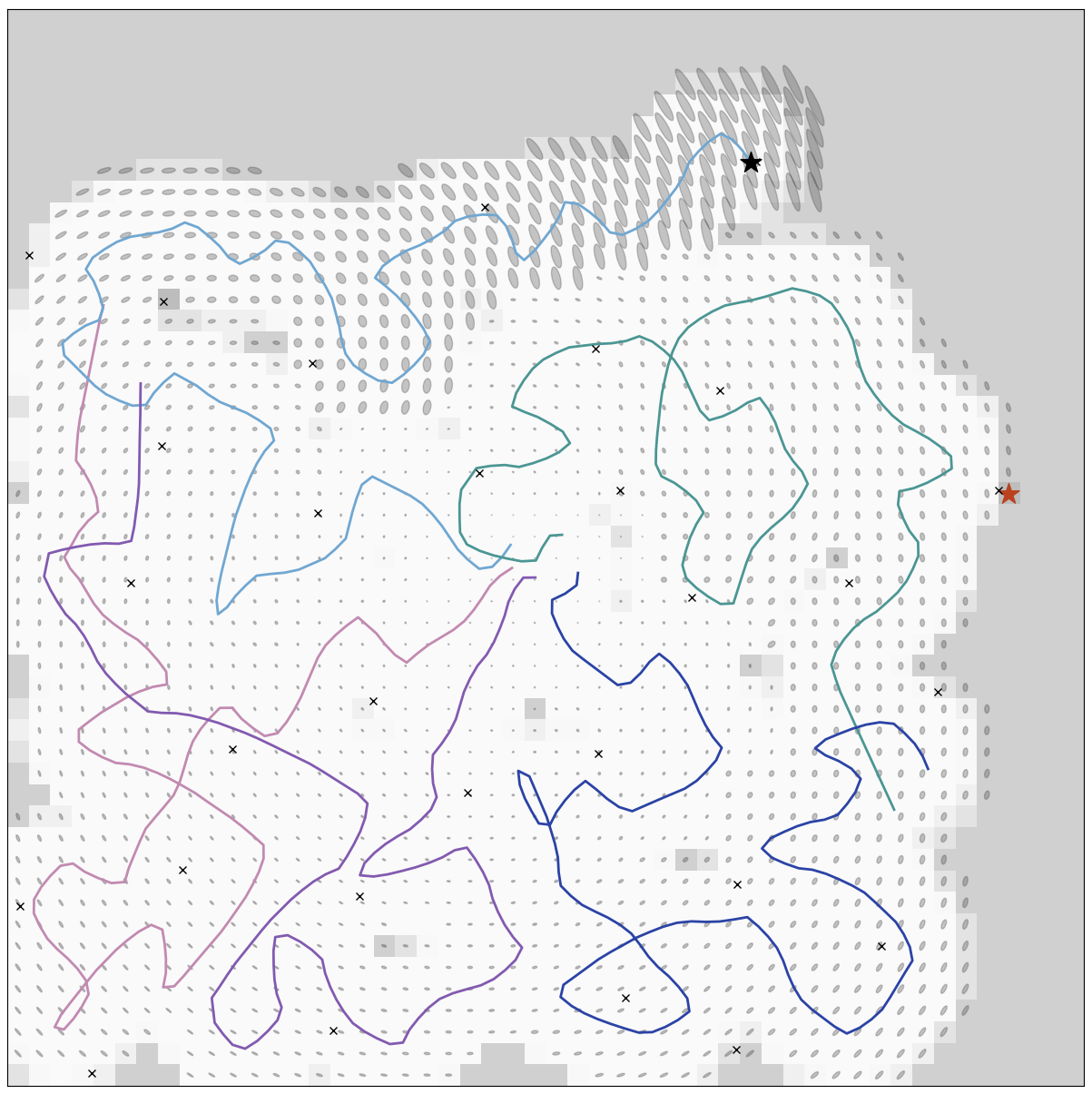}%
    }\hspace{0.2cm}
    \subfigure{%
    \includegraphics[width=0.4075\columnwidth]{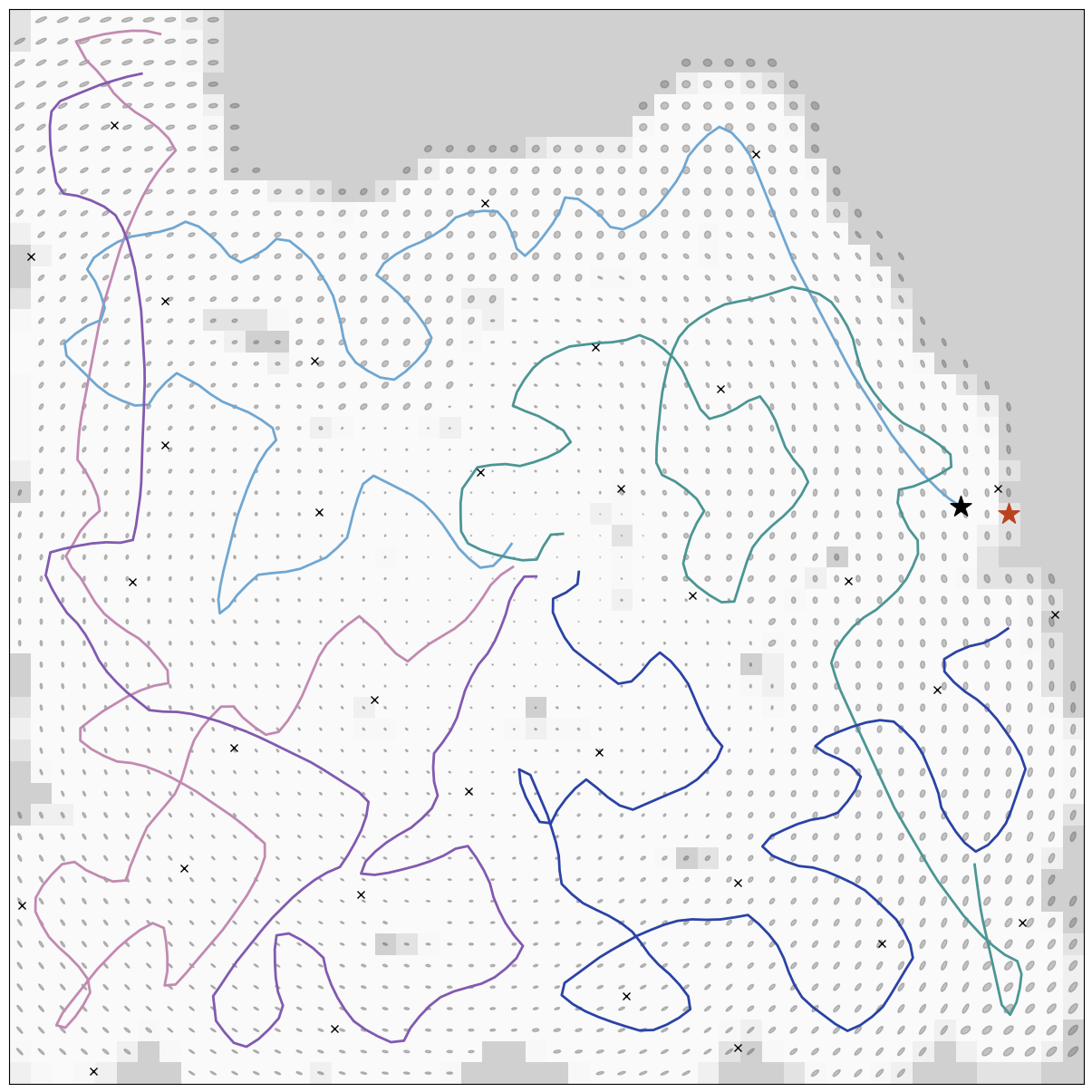}%
  } \vspace{-2mm}
  \caption{\textbf{Five robots navigating a 150m x 150m environment.} The active robot seeking its next target state is highlighted by a black star, while its chosen goal is marked with a red star. In the left map, the cyan robot revisits a previously explored landmark due to high map uncertainty (gray ellipses in each grid cell). After a loop closure in the right map, overall localization uncertainty decreases, prompting the robot to select the nearest unexplored frontier for its next move.}
\label{fig:obs_landmark}%
  \vspace{-5mm}
\end{figure}

\begin{figure}[t]
  \centering
  \subfigure{%
    \includegraphics[width=0.32\columnwidth]{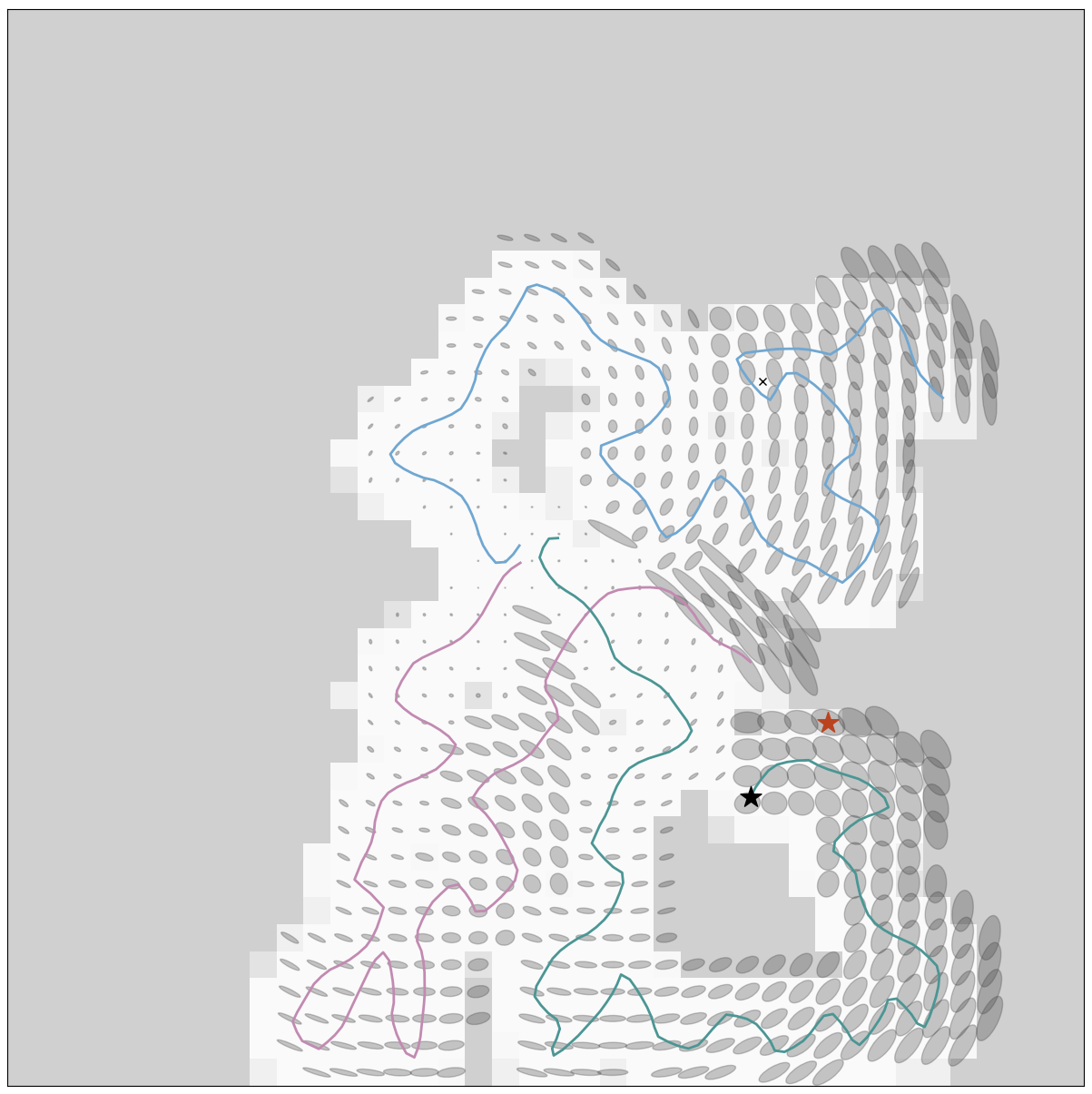}%
  }
  \subfigure{%
    \includegraphics[width=0.32\columnwidth]{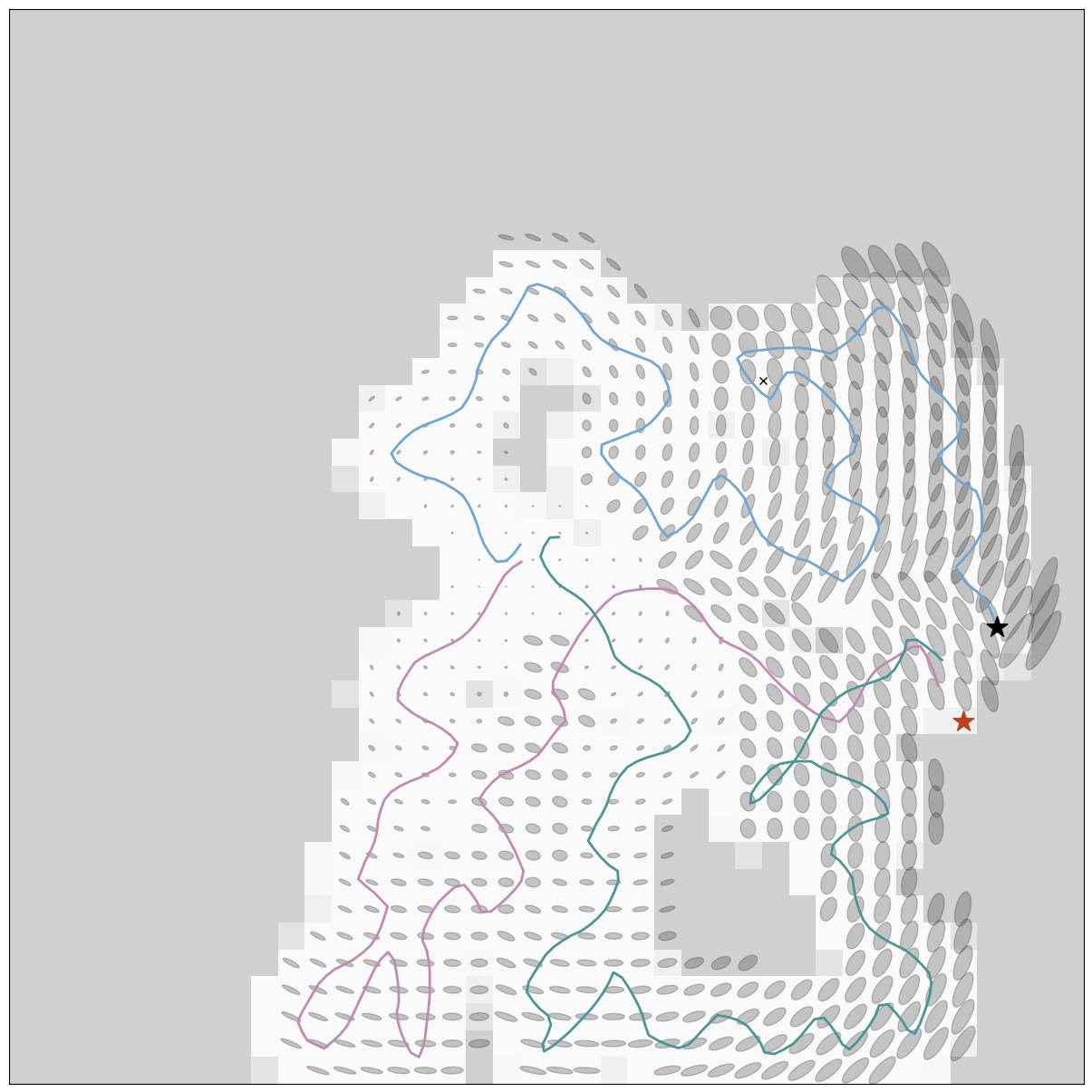}%
  }
  \subfigure{%
    \includegraphics[width=0.32\columnwidth]{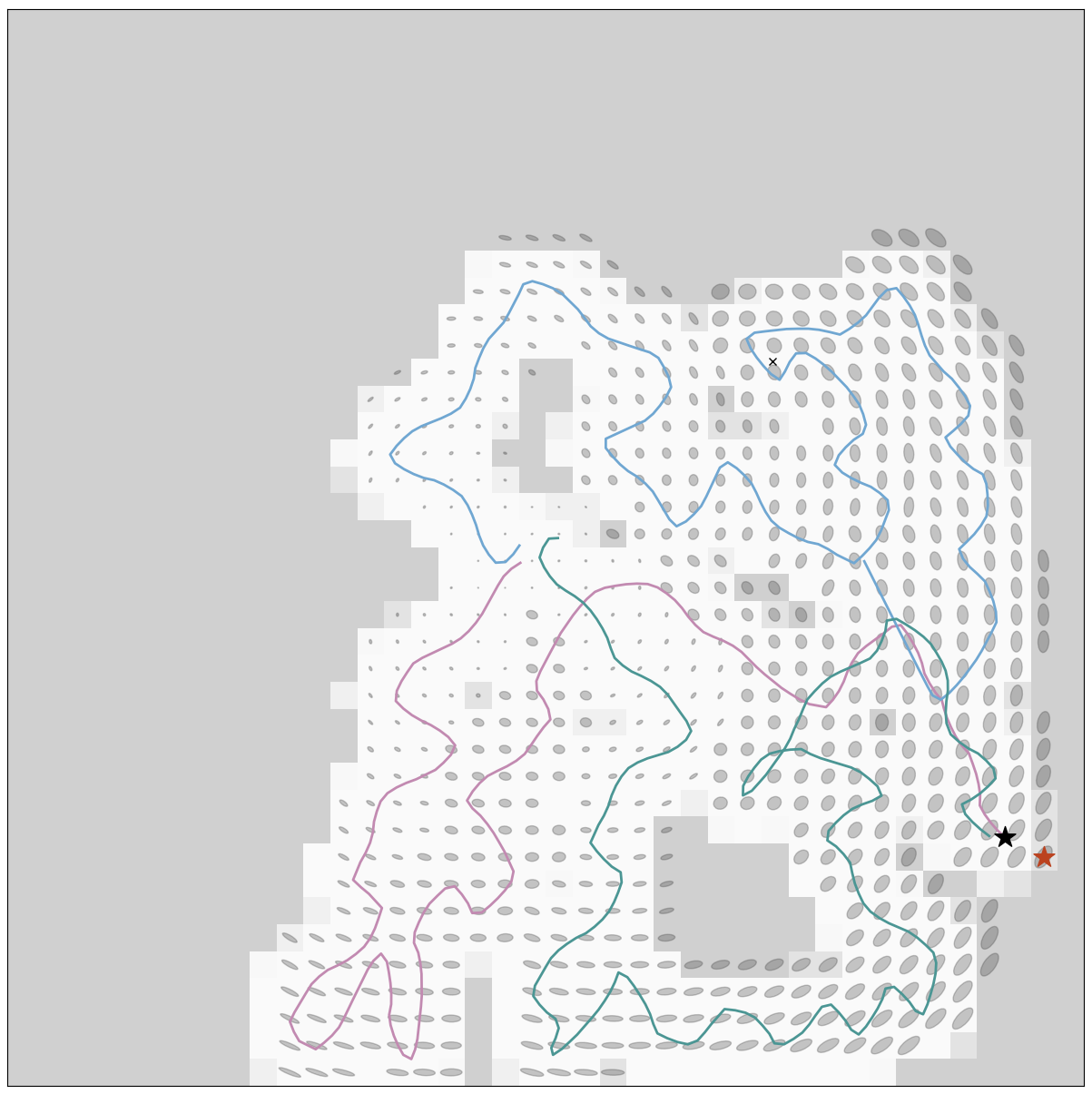}%
  } \vspace{-2mm}
  \caption{\textbf{Three robots explore a 200m x 200m environment.} Only the cyan robot has discovered a landmark, prompting the team of robots to strategically rendezvous with their teammates one by one.}
  \label{fig:obs_robot}
  \vspace{-8mm}
\end{figure}

Fig. \ref{fig:obs_landmark} and Fig. \ref{fig:obs_robot} present qualitative results of the proposed algorithm's EM\_3 configuration. 
In Fig. \ref{fig:obs_landmark}, five robots navigate themselves in a $150m \times 150m$ environment with 40 landmarks.
The cyan robot aggregates uncertainties (left) and decides to choose a new target goal state close to a landmark that has been visited by its teammate previously.
Three robots navigate a $200m \times 200m$ environment in Fig. \ref{fig:obs_robot}, with only one landmark observed thus far. After some time, the accumulated localization uncertainty is relatively high (left), prompting the green robot to decide to rendezvous with its pink neighbor. This strategic move leads to a reduction in uncertainty, thanks to the inter-robot loop closure (middle). Concurrently, the light blue robot also accumulates errors and opts to rendezvous with its teammates. Once all three robots meet each other (right), the loop is closed, and uncertainty is propagated and reduced.

\vspace{-2mm}
\section{Conclusions}\label{sec:conclusions}
\vspace{-1mm}
This paper introduces multi-robot exploration using expectation-maximization. We present an asynchronous exploration framework suitable for both centralized and decentralized robot teams, accounting for map uncertainty. The inclusion of rendezvous frontiers enhances the system's adaptability to environments with sparse features. The incorporation of a virtual map enables the estimation of future influence and interactions within the robot teams. The utilization of inter-robot and local maps heightens the robot's sensitivity to the accumulation of localization uncertainty in its trajectory. As a future direction, we aim to integrate this system with our previous decentralized multi-robot SLAM framework designed for underwater robot teams \cite{mcconnell2022draco} and subsequently validate the algorithm through experimentation on real robots.








\bibliographystyle{IEEEtran}
\bibliography{bib}

\begin{thebibliography}{10}
\providecommand{\url}[1]{#1}
\csname url@samestyle\endcsname
\providecommand{\newblock}{\relax}
\providecommand{\bibinfo}[2]{#2}
\providecommand{\BIBentrySTDinterwordspacing}{\spaceskip=0pt\relax}
\providecommand{\BIBentryALTinterwordstretchfactor}{4}
\providecommand{\BIBentryALTinterwordspacing}{\spaceskip=\fontdimen2\font plus
\BIBentryALTinterwordstretchfactor\fontdimen3\font minus
  \fontdimen4\font\relax}
\providecommand{\BIBforeignlanguage}[2]{{%
\expandafter\ifx\csname l@#1\endcsname\relax
\typeout{** WARNING: IEEEtran.bst: No hyphenation pattern has been}%
\typeout{** loaded for the language `#1'. Using the pattern for}%
\typeout{** the default language instead.}%
\else
\language=\csname l@#1\endcsname
\fi
#2}}
\providecommand{\BIBdecl}{\relax}
\BIBdecl

\bibitem{10075065}
J.~A. Placed, J.~Strader, H.~Carrillo, N.~Atanasov, V.~Indelman, L.~Carlone,
  and J.~A. Castellanos, ``A survey on active simultaneous localization and
  mapping: State of the art and new frontiers,'' \emph{IEEE Transactions on
  Robotics}, vol.~39, no.~3, pp. 1686--1705, 2023.

\bibitem{cadena2016past}
C.~Cadena, L.~Carlone, H.~Carrillo, Y.~Latif, D.~Scaramuzza, J.~Neira, I.~Reid,
  and J.~J. Leonard, ``Past, present, and future of simultaneous localization
  and mapping: Toward the robust-perception age,'' \emph{IEEE Transactions on
  Robotics}, vol.~32, no.~6, pp. 1309--1332, 2016.

\bibitem{burgard2005coordinated}
W.~Burgard, M.~Moors, C.~Stachniss, and F.~E. Schneider, ``Coordinated
  multi-robot exploration,'' \emph{IEEE Transactions on robotics}, vol.~21,
  no.~3, pp. 376--386, 2005.

\bibitem{fox2006distributed}
D.~Fox, J.~Ko, K.~Konolige, B.~Limketkai, D.~Schulz, and B.~Stewart,
  ``Distributed multirobot exploration and mapping,'' \emph{Proceedings of the
  IEEE}, vol.~94, no.~7, pp. 1325--1339, 2006.

\bibitem{jang2020multi}
D.~Jang, J.~Yoo, C.~Y. Son, D.~Kim, and H.~J. Kim, ``Multi-robot active sensing
  and environmental model learning with distributed gaussian process,''
  \emph{IEEE Robotics and Automation Letters}, vol.~5, no.~4, pp. 5905--5912,
  2020.

\bibitem{yu2021smmr}
J.~Yu, J.~Tong, Y.~Xu, Z.~Xu, H.~Dong, T.~Yang, and Y.~Wang, ``Smmr-explore:
  Submap-based multi-robot exploration system with multi-robot multi-target
  potential field exploration method,'' in \emph{IEEE International Conference
  on Robotics and Automation (ICRA)}, 2021, pp. 8779--8785.

\bibitem{atanasov2015decentralized}
N.~Atanasov, J.~Le~Ny, K.~Daniilidis, and G.~J. Pappas, ``Decentralized active
  information acquisition: Theory and application to multi-robot slam,'' in
  \emph{IEEE International Conference on Robotics and Automation (ICRA)}, 2015,
  pp. 4775--4782.

\bibitem{kontitsis2013multi}
M.~Kontitsis, E.~A. Theodorou, and E.~Todorov, ``Multi-robot active slam with
  relative entropy optimization,'' in \emph{American Control Conference (ACC)},
  2013, pp. 2757--2764.

\bibitem{corah2019communication}
M.~Corah, C.~O’Meadhra, K.~Goel, and N.~Michael, ``Communication-efficient
  planning and mapping for multi-robot exploration in large environments,''
  \emph{IEEE Robotics and Automation Letters}, vol.~4, no.~2, pp. 1715--1721,
  2019.

\bibitem{wang2020autonomous}
J.~Wang and B.~Englot, ``Autonomous exploration with
  expectation-maximization,'' in \emph{Robotics Research: The 18th
  International Symposium (ISRR)}.\hskip 1em plus 0.5em minus 0.4em\relax
  Springer, 2017, pp. 759--774.

\bibitem{charrow2014approximate}
B.~Charrow, V.~Kumar, and N.~Michael, ``Approximate representations for
  multi-robot control policies that maximize mutual information,''
  \emph{Autonomous Robots}, vol.~37, pp. 383--400, 2014.

\bibitem{colares2016next}
R.~G. Colares and L.~Chaimowicz, ``The next frontier: Combining information
  gain and distance cost for decentralized multi-robot exploration,'' in
  \emph{Proceedings of the 31st Annual ACM Symposium on Applied Computing},
  2016, pp. 268--274.

\bibitem{regev2016multi}
T.~Regev and V.~Indelman, ``Multi-robot decentralized belief space planning in
  unknown environments via efficient re-evaluation of impacted paths,'' in
  \emph{IEEE/RSJ International Conference on Intelligent Robots and Systems
  (IROS)}, 2016, pp. 5591--5598.

\bibitem{schlotfeldt2018anytime}
B.~Schlotfeldt, D.~Thakur, N.~Atanasov, V.~Kumar, and G.~J. Pappas, ``Anytime
  planning for decentralized multirobot active information gathering,''
  \emph{IEEE Robotics and Automation Letters}, vol.~3, no.~2, pp. 1025--1032,
  2018.

\bibitem{chen2022multi}
S.~Chen, L.~Zhao, S.~Huang, and Q.~Hao, ``Multi-robot active slam based on
  submap-joining for feature-based representation environments,'' in
  \emph{Australasian Conference on Robotics and Automation}, 2022.

\bibitem{lauri2017multi}
M.~Lauri, E.~Hein{\"a}nen, and S.~Frintrop, ``Multi-robot active information
  gathering with periodic communication,'' in \emph{IEEE International
  Conference on Robotics and Automation (ICRA)}, 2017, pp. 851--856.

\bibitem{freda20193d}
L.~Freda, M.~Gianni, F.~Pirri, A.~Gawel, R.~Dub{\'e}, R.~Siegwart, and
  C.~Cadena, ``3d multi-robot patrolling with a two-level coordination
  strategy,'' \emph{Autonomous Robots}, vol.~43, pp. 1747--1779, 2019.

\bibitem{kantaros2019asymptotically}
Y.~Kantaros, B.~Schlotfeldt, N.~Atanasov, and G.~J. Pappas, ``Asymptotically
  optimal planning for non-myopic multi-robot information gathering.'' in
  \emph{Robotics: Science and Systems}, 2019, pp. 22--26.

\bibitem{ginting2021chord}
M.~F. Ginting, K.~Otsu, J.~A. Edlund, J.~Gao, and A.-A. Agha-Mohammadi,
  ``Chord: Distributed data-sharing via hybrid ros 1 and 2 for multi-robot
  exploration of large-scale complex environments,'' \emph{IEEE Robotics and
  Automation Letters}, vol.~6, no.~3, pp. 5064--5071, 2021.

\bibitem{ye2022multi}
K.~Ye, S.~Dong, Q.~Fan, H.~Wang, L.~Yi, F.~Xia, J.~Wang, and B.~Chen,
  ``Multi-robot active mapping via neural bipartite graph matching,'' in
  \emph{IEEE/CVF Conference on Computer Vision and Pattern Recognition (CVPR)},
  2022, pp. 14\,839--14\,848.

\bibitem{tzes2023graph}
M.~Tzes, N.~Bousias, E.~Chatzipantazis, and G.~J. Pappas, ``Graph neural
  networks for multi-robot active information acquisition,'' in \emph{IEEE
  International Conference on Robotics and Automation (ICRA)}, 2023, pp.
  3497--3503.

\bibitem{ossenkopf2019long}
M.~Ossenkopf, G.~Castro, F.~Pessacg, K.~Geihs, and P.~De~Crist{\'o}foris,
  ``Long-horizon active slam system for multi-agent coordinated exploration,''
  in \emph{European Conference on Mobile Robots (ECMR)}, 2019, pp. 1--6.

\bibitem{indelman2018cooperative}
V.~Indelman, ``Cooperative multi-robot belief space planning for autonomous
  navigation in unknown environments,'' \emph{Autonomous Robots}, vol.~42, pp.
  353--373, 2018.

\bibitem{chen2020broadcast}
Y.~Chen, L.~Zhao, K.~M.~B. Lee, C.~Yoo, S.~Huang, and R.~Fitch, ``Broadcast
  your weaknesses: Cooperative active pose-graph slam for multiple robots,''
  \emph{IEEE Robotics and Automation Letters}, vol.~5, no.~2, pp. 2200--2207,
  2020.

\bibitem{kaess2012isam2}
M.~Kaess, H.~Johannsson, R.~Roberts, V.~Ila, J.~J. Leonard, and F.~Dellaert,
  ``isam2: Incremental smoothing and mapping using the bayes tree,'' \emph{The
  International Journal of Robotics Research}, vol.~31, no.~2, pp. 216--235,
  2012.

\bibitem{wang2022virtual}
J.~Wang, F.~Chen, Y.~Huang, J.~McConnell, T.~Shan, and B.~Englot, ``Virtual
  maps for autonomous exploration of cluttered underwater environments,''
  \emph{IEEE Journal of Oceanic Engineering}, vol.~47, no.~4, pp. 916--935,
  2022.

\bibitem{thrun2002probabilistic}
S.~Thrun, ``Probabilistic robotics,'' \emph{Communications of the ACM},
  vol.~45, no.~3, pp. 52--57, 2002.

\bibitem{kaess2009covariance}
M.~Kaess and F.~Dellaert, ``Covariance recovery from a square root information
  matrix for data association,'' \emph{Robotics and autonomous systems},
  vol.~57, no.~12, pp. 1198--1210, 2009.

\bibitem{fan2020improved}
X.~Fan, Y.~Guo, H.~Liu, B.~Wei, and W.~Lyu, ``Improved artificial potential
  field method applied for {AUV} path planning,'' \emph{Mathematical Problems
  in Engineering}, vol. 2020, pp. 1--21, 2020.

\bibitem{mcconnell2022draco}
J.~McConnell, Y.~Huang, P.~Szenher, I.~Collado-Gonzalez, and B.~Englot,
  ``D{RAC}o-{SLAM}: {D}istributed robust acoustic communication-efficient
  {SLAM} for imaging sonar equipped underwater robot teams,'' in \emph{IEEE/RSJ
  International Conference on Intelligent Robots and Systems (IROS)}, 2022, pp.
  8457--8464.

\end{thebibliography}

\end{document}